\documentclass{llncs}
\usepackage{amsmath,amssymb}
\usepackage{graphics}
\usepackage{tikz}
\usetikzlibrary{arrows,shapes,positioning}
\usepackage{caption}
\usepackage[linesnumbered,ruled,vlined]{algorithm2e}
\usepackage{subfig}
\usepackage{enumitem}
\usepackage{diagbox}
\usepackage{hhline}
\usepackage{array,multirow,graphicx}

\usepackage{float}

\title{Tunable Online MUS/MSS Enumeration}

\author{Jaroslav Bend\'ik \and Nikola Bene\v s \and Ivana \v Cern\'a \and Ji\v r\'i Barnat}

\institute{Faculty of Informatics, Masaryk University, Brno, Czech Republic\\
\email{\{xbendik,xbenes3,cerna,barnat\}@fi.muni.cz}}

\begin{document}
\maketitle

\begin{abstract}
In various areas of computer science, the problem of dealing with a~set of
constraints arises. If the set of constraints is unsatisfiable, one may ask
for a~minimal description of the reason for this unsatisifiability. Minimal
unsatisfiable subsets (MUSes) and maximal satisfiable subsets (MSSes) are two
kinds of such minimal descriptions. The goal of this work is the enumeration
of MUSes and MSSes for a~given constraint system. As such full enumeration
may be intractable in general, we focus on building an online algorithm,
which produces MUSes/MSSes in an on-the-fly manner as soon as they are
discovered.  The problem has been studied before even in its online version.
However, our algorithm uses a~novel approach that is able to outperform the
current state-of-the art algorithms for online MUS/MSS enumeration. Moreover,
the performance of our algorithm can be adjusted using tunable parameters.
We evaluate the algorithm on a~set of benchmarks.
\end{abstract}

\section{Introduction}

In various areas of computer science, such as constraint processing,
requirements analysis, and model checking, the following problem often arises.
We are given a~set of constraints and are asked whether the set of constraints
is feasible, i.e.~whether all the constraints are satisfiable together.
In requirements analysis, the constraints represent the requirements on a~given
system, usually described as formulae of a~suitable logic, and the feasibility
question is in fact the question whether all the requirements can actually be
implemented at once. In some model checking systems, such as those using the
counterexample-guided abstraction refinement (CEGAR) workflow, an infeasible
constraint system may arise as a result of the abstraction's overapproximation.
In such cases where the set of constraints is infeasible, we might want to
explore the reasons of infeasability. There are basically two approaches that
can be used here. One is to try to extract a~single piece of information
explaining the infeasibility, such as a minimal unsatisfiable subset (MUS)
or dually a~maximal satisfiable subset (MSS) of the constraints. The other
option is to try to enumerate all, or at least as many as possible, of these
sets. In this work, we focus on the second approach. Enumerating multiple
MUSes is sometimes desirable: in requirements analysis, this gives better
insight into the inconsistencies among requirements; in CEGAR-based model
checking more MUSes lead to a~better refinement that can reduce the complexity
of the whole procedure~\cite{reveal}.

The enumeration of all MUSes or MSSes is generally intractable due to the
potentially exponential number of results.  It thus makes sense to study
algorithms that are able to provide at least some of those within a~given time
limit. An even better option is to have an algorithm that produces MUSes or
MSSes in an on-the-fly manner as soon as they are discovered. It is the goal of
this paper to describe such an algorithm.

\subsection{Related Work}

The list of existing work that focuses on enumerating multiple MUSes is short
as most of the related work focused just on an extraction of a single MUS or
even a~non-minimal unsatisfiable subset. For example
all of \cite{DBLP:conf/dac/OhMASM04,bruni,zhang2003extracting}
uses information from a satisfiability solver
to obtain an unsatisfiable subset but they do not guarantee its minimality.
Moreover, the majority of the algorithms which enumerate all MUSes have been
developed for specific constraint domains, mainly for Boolean satisfiability
problems.


\subsubsection{Explicit checking}
The first algorithm for enumerating all MUSes we are aware of was developed by
Hou~\cite{hou} in the field of diagnosis and is built on explicit enumeration
of every subset of the unsatisfiable constraint system. It checks every subset
for satisfiability, starting from the complete constraint set and branching in
a tree-like structure. The authors presented some pruning
rules to skip irrelevant branches and avoid unnecessary work. Further
improvements to this approach were made by Han and Lee~\cite{han} and also by
de la Banda et. al.~\cite{banda}.


\subsubsection{CAMUS}
A state-of-the-art algorithm for enumerating all MUSes called CAMUS by Liffiton
and Sakallah~\cite{camus} is based on the relationship between MUSes and the
so-called minimal correction sets (MCSes), which was independently pointed out
by \cite{daa,birnbaum,DBLP:conf/ijcai/LiffitonMPS05}. This relationship states
that $M \subseteq C$ is a MUS of $C$ if and only if it is an irreducible
hitting set of MCS($C$). CAMUS works in two phases, first it computes all MCSes
of the given constraint set, and then it finds all MUSes by computing all the
irreducible hitting sets of these MCSes.


A significant shortcoming of CAMUS is that the first phase can be intractable
as the number of MCSes may be exponential in the size of the instance and all
MCSes must be enumerated before any MUS can be produced. This makes CAMUS
unsuitable for many applications which require only a few MUSes but want to
get them quickly.
Note that CAMUS is able to enumerate MSSes, as they are simply the complements
of MCSes.

\subsubsection{MARCO}
The desire to enumerate at least some MUSes even in the generally intractable
cases led to the development of two independent but nearly identical algorithms:
MARCO~\cite{marco-original} and eMUS~\cite{emus}. Both algorithms were later
joined and presented in~\cite{marco} under the name of MARCO.
MARCO is able to produce individual MUSes during its execution and it does it
in a relatively steady rate. To obtain each single MUS, MARCO first
finds a subset $U$ whose satisfiability is not known yet, checks it for
satisfiability and if it is unsatisfiable, it is ``shrunk'' to a MUS.
In the case that $U$ is satisfiable, it is in a dual manner expanded into a~MSS.
The algorithm can be supplied with any appropriate shrink and expansion
procedures; this makes MARCO applicable to any constraint satisfaction domain
in general.

CAMUS and MARCO were experimentally compared in~\cite{marco} and the former has shown
to be faster in enumerating all MUSes in the tractable cases.  However, in the
intractable cases, MARCO was able to provide at least some MUSes while CAMUS
often provided none. One another algorithm, the Dualize and Advance (DAA) by
Bailey and Stuckey~\cite{daa} was also evaluated in these experiments. DAA is
also based on the relationship between MCSes and MUSes and can produce both
MUSes and MSSes during its execution; however, it has shown to be substantially
slower than CAMUS in the case of complete MUSes enumeration and also slower
than MARCO in the partial enumeration.

\subsection{Our Contribution}

In this paper, we present our own algorithm for online enumeration of MUSes and
MSSes in general constraint satisfaction domains that is able to outperform
the current state-of-the-art MARCO algorithm.  The core of the algorithm is
based on a~novel binary-search-based approach. Similarly to MARCO, the
algorithm is able to directly employ arbitrary shrinking and expanding
procedures. Moreover, our algorithm contains certain parameters that govern in
which cases the shrinking and expanding procedures are to be used.  We evaluate
our algorithm on a~variety of benchmarks that show that the algorithm indeed
outperforms MARCO.

\subsection{Outline of The Paper}
In Section~\ref{sec:prelim} we state the problem we are solving in a~formal
way, defining all the necessary notions. In Section~\ref{sec:algo} we describe
the algorithm in an incremental way, starting with the basic schema of MUS/MSS
computation and gradually explaining the main ideas of our algorithm.
Section~\ref{sec:exp} provides an experimental evaluation on a variety of
benchmarks, comparing our algorithm against MARCO. The paper is concluded in
Section~\ref{sec:concl}.

\section{Preliminaries}\label{sec:prelim}

Our goal is to deal with arbitrary constraint satisfaction system. The input is
given as a~finite set of constraints $C = \{c_1, c_2, \ldots, c_n\}$
with the property that each subset of $C$ is either \emph{satisfiable} or
\emph{unsatisfiable}. The definition of satisfiability may vary in different
constraint domains, we only assume that if $X \subseteq C$ is satisfiable,
then all subsets of $X$ are also satisfiable.
The subsets of interest are defined in the following.

\begin{definition}[MSS, MCS, MUS]
Let $C$ be a~finite set of constraints and let $N \subseteq C$.
$N$ is a~\emph{maximal satisfiable subset} (MSS) of $C$ if $N$ is satisfiable
and $\forall c \in C \setminus N : N \cup \{c\}$ is unsatisfiable.
$N$ is a~\emph{minimal correction set} (MCS) of $C$ if $C \setminus N$ is a~MSS
of $C$.
$N$ is a~\emph{minimal unsatisfiable subset} (MUS) of $C$ if $N$ is
unsatisfiable and $\forall c \in N : N \setminus \{c\}$ is satisfiable.
\end{definition}

Note that the maximality concept used here is set maximality, not maximum
cardinality as in the MaxSAT problem. This means that there can be multiple
MSSes with different cardinality.
We use MUS($C$), MCS($C$), and MSS($C$) to denote the set of all MUSes, MCSes,
and MSSes of $C$, respectively.
The formulation of our problem is the following: Given a~finite set of
constraints $C$, enumerate (all or at least as many as possible) members of
MUS($C$) and MSS($C$). Note that due to the complementarity of MSS and MCS,
this also enumerates all MCS($C$).

To describe the ideas of our algorithm and illustrate its usage, we shall use
Boolean satisfiability constraints in the following. In the examples,
each of the constraints $c_i$ is going to be a~clause (a~disjunction of
literals). The whole set of constraints can be then seen as a~Boolean formula
in conjunctive normal form.

\begin{example}\label{ex:small}
We illustrate the concepts on a small example. Assume that we are given a set
$C$ of four Boolean satisfiability constraints $c_1 = a$, $c_2 = \neg a$, $c_3
= b$, and $c_4 = \neg a \vee \neg b$.  Clearly, the whole set is unsatisfiable
as the first two constraints are negations of each other. There are two MUSes:
$\{c_1,c_2\}$, $\{c_1,c_3,c_4\}$, three MSSes: $\{c_1,c_4\}$, $\{c_1,c_3\}$,
$\{c_2,c_3,c_4\}$ and three MCSes: $\{c_2,c_3\}$, $\{c_2,c_4\}$, $\{c_1\}$.
\end{example}

The \emph{powerset} of $C$, i.e.~the set of all its subsets, forms a lattice
ordered via subset inclusion and denoted by $\mathcal{P}(C)$. 
In our algorithm we are going to deal with the so-called \emph{chains} of
the powerset and deal with local MUSes and MSSes, defined as follows.

\begin{definition}
Let $C$ be a~finite set of constraints. The sequence
$K = \langle N_1, \ldots N_i \rangle$ is a \emph{chain} in $\mathcal{P}(C)$ if
$\forall j : N_j \in \mathcal{P}(C)$ and 
$N_1 \subset N_2 \subset \cdots \subset N_i$. 
We say that $N_k$ is a~\emph{local MUS} of $K$ if $N_k$ is unsatisfiable and 
$\forall j < k : N_j$ is satisfiable.
Similarly, we say that $N_k$ is a~\emph{local MSS} of $K$ if $N_k$ is satisfiable
and $\forall j > k : N_j$ is unsatisfiable.
\end{definition}
Note that there is no local MUS if all subsets on the chain are satisfiable,
and there is no local MSS if all subsets on the chain are unsatisfiable.


\section{Algorithm}\label{sec:algo}

In this section, we gradually present an online MUS/MSS enumeration algorithm.
Consider first a~naive enumeration algorithm that would explicitly check each
subset of $C$ for satisfiability, split the subsets of $C$ into satisfiable and
unsatisfiable subsets, and choose the maximal and minimal subsets of the two
groups, respectively. The main disadvantage of this approach is the large
number of satisifiability checks. Checking a~given subset of $C$ for
satisfiability is usually an expensive task and the naive solution makes an
exponentially many of these checks which makes it unusable.

Note that the problem of MUS enumeration contains the solution to the problem
of satisfiability of all subsets of $C$ as each unsatisfiable subset of $C$
is a~superset of some MUS. This means that every algorithm that solves the
problem of MUS enumeration has to make several satisfiability checks during its
execution. These checks are usually done employing an external satisfiability
solver. Clearly, the number of such external calls corresponds with the
efficiency of the algorithm. It is therefore our goal to minimise the number
of calls to the solver.

\subsection{Basic Schema}

Recall that the elements of $\mathcal{P}(C)$ are partially ordered via subset
inclusion and each element is either satisfiable or unsatisfiable. The key
assumption on the constraint domain, as declared above, is that the partial
ordering of subsets is preserved by the satisfiability of these subsets.
If we thus find an unsatisfiable subset $N_u$ of $C$ then all supersets of
$N_u$ are also unsatisfiable; dually, if we find a~satisfiable subset $N_s$
of $C$ then all subsets of $N_s$ are also satisfiable.
Moreover, none of the supersets of $N_u$ can be a MUS and none of the subsets
of $N_s$ can be a MSS. In the following text we refer to this property as to
the \emph{monotonicity} of $\mathcal{P}(C)$, and to the elements of
$\mathcal{P}(C)$ as to \emph{nodes}.

Our basic algorithm is described in pseudocode as Algorithm~\ref{alg-basic}.
The algorithm consists of two phases. In the first phase it determines
the satisfiability of all nodes and extracts from $\mathcal{P}(C)$ a set of MSS
\emph{candidates} $MSS_{can}$ and a set of MUS candidates $MUS_{can}$ ensuring
that $MSS(C) \subseteq MSS_{can}$ and $MUS(C) \subseteq MUS_{can}$. In the
second phase it reduces $MSS_{can}$ to $MSS(C)$ and $MUS_{can}$ to $MUS(C)$.

During the execution of the first phase the algorithm maintains a
classification of nodes; each node can be either \emph{unexplored} or
\emph{explored} and some of the explored nodes can belong to $MSS_{can}$ or to
$MUS_{can}$. \emph{Explored} nodes are those, whose satisfiability the
algorithm already knows and \emph{unexplored} are the others. The algorithm
stores the unexplored nodes in the set $Unex$ which initially contains all
nodes from $\mathcal{P}(C)$. The first phase is iterative, the algorithm in
each iteration selects some unexplored nodes $Nodes$, determines their
satisfiability using an external satisfiability solver, and exploits the
monotonicity of $\mathcal{P}(C)$ to deduce satisfiability of some other
unexplored nodes. At the end of each iteration the algorithm updates the set
$Unex$ by removing from it the nodes whose satisfiability was decided in this
iteration. 
Based on its satisfiability, every node from the set $Nodes$ is added either
into $MSS_{can}$ or $MUS_{can}$.

In the pseudocode, we use $Sup(N)$ to denote the set of all unexplored
supersets of $N$ including $N$ and $Sub(N)$ to denote the the set of all
unexplored subsets of $N$ including $N$. The notation $\overline{Sup(N)}$,
$\overline{Sub(N)}$ is used to denote the complements of $Sup(N)$ and $Sub(N)$.

%

\begin{algorithm}[t]
\SetKwInOut{Input}{input}\SetKwInOut{Output}{output}
\DontPrintSemicolon

 $Unex \gets \mathcal{P}(C)$\;
 $MSS_{can}, MUS_{can} \gets \emptyset$\;
 \While{$Unex$ is not empty}{
	$Nodes \gets$ some unexplored nodes\;
	\For{each $N \in Nodes$}{
		\eIf{$N$ is satisfiable}{
			$MSS_{can} \gets MSS_{can} \cup \{N\}$\;
			$Unex \gets Unex \cap \overline{Sub(N)}$\;
		}{
			$MUS_{can} \gets MUS_{can} \cup \{N\}$\; 
			$Unex \gets Unex \cap \overline{Sup(N)}$\;
		}
	}
 }
 extract MSSes from $MSS_{can}$\;
 extract MUSes from $MUS_{can}$\; 
 \caption{The basic schema of our algorithm}
 \label{alg-basic}
\end{algorithm}

Clearly, the schema converges as the set of unexplored nodes decreases its size
in every iteration. The schema also ensures that after the last iteration it
holds that $MUS(C) \subseteq MUS_{can}$ and $MSS(C) \subseteq MSS_{can}$. This
is directly implied by the monotonicity of $\mathcal{P}(C)$ as no node whose
satisfiability was deduced can be a MSS and dually no node whose
unsatisfiability was deduced can be a MUS.

In the second phase our algorithm extracts all MUSes and MSSes from $MUS_{can}$ and $MSS_{can}$.
Both these extractions can be done by any algorithm that extracts the highest
and the lowest elements from any partially ordered set. A trivial algorithm can
just test each pair of elements for the subset inclusion and remove the
undesirable elements, which can be done in time polynomial to the number of
constraints in $C$ and the size of the sets of candidates. We assume that this
part of our algorithm is not as expensive as the rest of it, especially when
each check for a satisfiability of a set of constraints may require solving an
NP-hard problem. We therefore omit the discussion of the second phase in the
following and focus solely on the way the set $Nodes$ is chosen in each
iteration and the way the unexplored nodes are managed.


\subsection{Symbolic Representation of Nodes}

Our algorithm highly depends on an efficient management of nodes. In
particular it needs to reclassify some nodes from unexplored to explored
and build chains from the unexplored nodes. Probably the simplest way of
managing nodes would be their explicit enumeration; however, there are
exponentially many subsets of $C = \{c_1, \cdots , c_n\}$ and their explicit
enumeration is thus intractable for large instances.
We thus use a symbolic representation of nodes instead.

We use the fact that the powerset lattice $\mathcal P(C)$ can be seen and
manipulated as a~Boolean algebra. We thus encode the set of constraints
$C = \{c_1, \ldots, c_n\}$ using a~set of Boolean variables
$X = \{x_1, \ldots, x_n\}$. Each subset of $C$ (i.e.~each node in our
algorithm) is then represented by a~valuation of the variables of $X$. 
This allows us to represent sets of nodes using Boolean formulae over $X$.
We use $f(Nodes)$ to denote the Boolean formula representing the set $Nodes$
in the following.



As an example, consider a~set of constraints $C = \{c_1, c_2, c_3\}$ and let
$Nodes = \{\{c_1\}, \{c_1,c_2\}, \{c_1,c_3\} \}$ be a set of three nodes.
Using the Boolean variables representation of $C$, we can encode the set
$Nodes$ using the Boolean formula $f(Nodes) = x_1 \wedge (\neg x_2 \vee \neg x_3)$.

The advantage of this representation is that we can efficiently perform set
operations over sets of nodes. The union of two sets of nodes $NodesA, NodesB$
is carried out as a disjunction and their intersection as a conjunction.
To get an arbitrary node from a~given set, say $Unex$, we use an external SAT
solver (more details in the next subsection). Note that this means that our
algorithm employs two external solvers: One is the constraint satisfaction
solver that decides satisfiability of the nodes, one is the SAT solver that 
works with our Boolean description of the constraint set and is employed
to produce unexplored nodes. To clearly distinguish between these two we shall
in the following use the phrases ``constraint solver'' and ``SAT solver''
rigorously.

\subsection{Unexplored Nodes Selection} 

Let us henceforth denote one specific call to the constraint solver as a~\emph{check}.
We now clarify which nodes our algorithm chooses in each of its iterations to
be \emph{checked} and which nodes it adds into the sets of candidates on MUSes
and MSSes. We also extend the basic schema which was presented as
Algorithm~\ref{alg-basic}.
We want to minimise the ratio of performed checks to the number of nodes in
$\mathcal{P}(C)$. 
Every algorithm for solving the problem of MUSes enumeration has to perform at
least as many checks as there are MUSes, so this ratio can never be zero.
Also, it is impossible to achieve the ratio with a~minimal value without knowing
which nodes are satisfiable and which are not and this information is not
a part of the input of our algorithm. Instead of minimising this overall ratio,
our algorithm tends to minimise this ratio locally in each of its iterations. 


In order to select the nodes which are checked in one specific iteration, our
algorithm at first constructs an \emph{unexplored chain}. An \emph{unexplored
chain} is a chain $K = \langle N_1, \ldots , N_k \rangle$ that contains only
unexplored nodes and that cannot be extended by adding another unexplored
nodes to its ends, i.e.~$N_1$ has no unexplored subset and $N_k$ has no
unexplored superset. The monotonicity of $\mathcal{P}(C)$ implies that either (i)
all nodes of $K$ are satisfiable, (ii) all nodes of $K$ are unsatisfiable, or
(iii) $K$ has a local MSS and a local MUS, i.e.~there is some $j$ such that
$\forall 0 \leq i \leq j$ : $N_i$ is satisfiable and $\forall k \geq l > j$ :
$N_l$ is unsatisfiable. This allows us to employ binary search to find such
$j$ performing only logarithmically many checks in the length of the chain.
Let us analyse the three possible cases:


\begin{enumerate}
	\item[(i)] all nodes of $K$ are satisfiable, hence our algorithm deduces that all proper subsets of $N_k$ are satisfiable and none of them can be a MSS, and it marks $N_k$ as a MSS candidate;
	\item[(ii)] all nodes of $K$ are unsatisfiable, hence our algorithm deduces that all proper supersets of $N_1$ are unsatisfiable and none of them can be a MUS, and it marks $N_1$ as a MUS candidate; or
	\item[(iii)] $N_j$ is the local MSS of $K$ and $N_{j+1}$ is its local MUS, hence our algorithm deduces that all proper subsets of $N_j$ are satisfiable, all proper supersets of $N_{j+1}$ are unsatisfiable, and it marks $N_j$ as a~MSS candidate and $N_{j+1}$ as a~MUS candidate.
\end{enumerate}

Algorithm~\ref{alg-extended} shows the extended schema of our algorithm which
implements the above method for choosing nodes to be checked.  At the beginning
of each iteration the algorithm finds an unexplored chain $K$ which is
subsequently processed by the $processChain$ method. This method finds the
local MUS and local MSS of $K$ (possibly only one of those) using binary
search and returns them. 

\begin{algorithm}[t]
 \SetKwInOut{Input}{input}\SetKwInOut{Output}{output}
\SetKwFunction{pC}{processChain}
\DontPrintSemicolon

 $Unex \gets \mathcal{P}(C)$\;
 $MSS_{can}, MUS_{can} \gets \emptyset$\;
 \While{$Unex$ is not empty}{
 	$K \gets$ some unexplored chain\;
 	$Nodes \gets \pC(K)$\;
	\For{each $N \in Nodes$}{
		\eIf{$N$ is satisfiable}{
			$MSS_{can} \gets MSS_{can} \cup \{N\}$\;
			$Unex \gets Unex \cap \overline{Sub(N)}$\;
		}{
			$MUS_{can} \gets MUS_{can} \cup \{N\}$\; 
			$Unex \gets Unex \cap \overline{Sup(N)}$\;
		}
	}
 }
 extract MSSes from $MSS_{can}$\;
 extract MUSes from $MUS_{can}$\; 

 \caption{The extended schema of our algorithm}
 \label{alg-extended}
\end{algorithm}

To construct an unexplored chain, our algorithm first finds a pair of
unexplored nodes $(N_1, N_k)$ such that $N_1 \subseteq N_k$ and then builds a
chain $\langle N_1, N_2, \ldots,$ $ N_{k-1}, N_k \rangle$ by connecting these two
nodes. The intermediate nodes $N_2, \ldots, N_{k-1}$ are obtained by adding one
by one the constraints from $N_k \setminus N_1$ to the node $N_1$. We refer to
each such pair of unexplored nodes $(N_1, N_k)$ that are the end nodes of some
unexplored chain as to an \emph{unexplored couple}.

In order to find an unexplored couple our algorithm asks for a~member of
$Unex$ by employing the SAT solver (by asking for a~model of the formula
$f(Unex)$). Besides the capability of finding an
arbitrary member of $Unex$, we require the following capability:
For a~given member $N_p \in Unex$, the SAT solver should be able to produce
a~\emph{minimal} $N_q \in Unex$ such that $N_q \subseteq N_p$,
where \emph{minimal} means that there is no other $N_r$ with $N_r \subset N_q$.
Similarly, we require the SAT solver to be able to produce \emph{maximal}
such $N_q$.
One of the SAT solvers that satisfies our requirements is miniSAT \cite{minisat}
that allows the user to fix values of some variables and to select a default polarity of
variables at decision points during solving. To obtain a minimal $N_q$ which is
a subset of $N_p$, we set the default polarity of variables to False and fix
the truth assignment to the variables that have been assigned False in $N_p$.
Similarly for the maximal case.

We now describe two approaches of obtaining unexplored couples, assuming that
we employ a~SAT solver satisfying the above requirements.

\medskip
\textbf{Basic approach} The \emph{Basic approach} consists of two calls to the
SAT solver. The first call asks the SAT solver for an arbitrary minimal member
of $Unex$. If nothing is returned then there are no more unexplored nodes.
Otherwise we obtain a~node $N_k$ which is minimal in $Unex$. We then ask the
SAT solver for a maximal node $N_l \in Unex$ such that $N_l$ is a superset of
$N_k$. The pair $(N_k, N_l)$ is then the new unexplored couple. 

\medskip
\textbf{Pivot based approach} 
Supposing that the SAT solver works deterministically, a series of calls for
maximal (minimal) nodes of $Unex$ may return nodes from some local part of the search space
that may lead to construction of unnecessarily short chains. In order to
alleviate this disadvantage of the Basic approach we propose to first choose
a \emph{pivot} $N_p$, an unexplored node which may be neither maximal nor
minimal and which should be chosen somehow randomly. As the next step this
approach asks the SAT solver for a minimal node $N_l$ such that $N_k \subseteq
N_p$ and for a maximal node $N_l$ such that $N_p \subseteq N_l$. The new
unexplored couple is then $(N_k, N_l)$. The randomness in choosing the node
$N_p$ is expected to ensure that we hit a~part of $Unex$ with large chains.

To get the pivot, we create a random partial valuation by randomly fixing
values of some variables and ask the SAT solver for a node that complies with
this partial valuation. If the solver returns a~node, we use it as the pivot.
Clearly, giving the SAT solver a partial valuation may make it fail to find
a~node despite the fact that there still are some. Therefore, if the solver
return nothing, we try to get the unexplored couple using the Basic approach.

\subsection{Online MUS/MSS Enumeration}

The algorithm as presented until now is only able to provide MUSes and MSSes in
the second phase, after it finished exploring all the nodes. 
We now describe the last piece of our final algorithm, namely the way of 
producing MUSes and MSSes during the execution of the first phase.
To do so, we need to employ two procedures:
The \emph{shrink} procedure is an~arbitrary method that can turn a
unsatisfiable node $N_u$ into a MUS. Dually, the \emph{grow} procedure is
a~method that can turn a satisfiable node into MSS $N_s$. A simple variant of
these two procedures is shown in Algorithms~\ref{alg:shrink}
and~\ref{alg:grow}. The simple shrink (grow) method iteratively attempts to remove
(add) constraints from $N_u$ ($N_s$), checking each new set for satisfiability
and keeping any changes that leave the set unsatisfiable (satisfiable). These
simple variants serve just as illustrations, there are known efficient
implementations of both shrink and grow for specific constraint domains;
as an example see MUSer2~\cite{muser} which implements the shrink method for
Boolean constraints systems.

\begin{figure}[t]
\begin{minipage}{0.5\textwidth}
\begin{algorithm}[H]
  {\normalsize
  \SetKwInOut{Input}{input}\SetKwInOut{Output}{output}
\DontPrintSemicolon
\For{$c \in N_u$}{
	\If{$N_u \setminus \{c\}$ is unsatisfiable}{
		$N_u \gets N_u \setminus \{c\}$
	}
}
\Return{$N_u$}
  \caption{shrink($C,N_u$)}
  \label{alg:shrink}}
\end{algorithm}
\end{minipage}
\vrule\hspace{0.1em}
\begin{minipage}{0.48\textwidth}
\begin{algorithm}[H]
  {\normalsize

\SetKwInOut{Input}{input}\SetKwInOut{Output}{output}
\DontPrintSemicolon

\For{$c \in C \setminus N_s$}{
	\If{$N_s \cup \{c\}$ is satisfiable}{
		$N_s \gets N_s \cup \{c\}$
 	}
}
\Return{$N_s$}  
  \caption{grow($C,N_s$)}
  \label{alg:grow}}
\end{algorithm}
\end{minipage}
\end{figure}

\begin{algorithm}[t]

\SetKwInOut{Input}{input}\SetKwInOut{Output}{output}
\SetKwFunction{yMU}{yieldMUS}
\SetKwFunction{yMS}{yieldMSS}
\DontPrintSemicolon
find local MSS $N_s$ and MUS $N_u$ of $K$ using binary search\;
\If{$u < S(|K|)$}{
	$N_u \gets shrink(N_u)$\;
    $\yMU(N_u)$
}
\If{$s > |K| - G(|K|)$}{
	$N_s \gets grow(N_s)$\;
	$\yMS(N_s)$
}

\Return{$\{N_u,N_s\}$} \tcp*{Note that $N_u$ or $N_s$ may not exist}
 \caption{processChain($C,K = \langle N_1, \ldots , N_k \rangle$)}
 \label{alg-process-chain}
\end{algorithm}

Recall that as a result of a processing a single chain $K$, our algorithm finds
either a local MUS $N_u$, or a local MSS $N_s$, or both of them. To get a MUS
(MSS) we propose to employ the shrink (grow) method on this local MUS (MSS).
However, performing shrink (grow) on each local MUS (MSS) can be quite
expensive and can significantly slow down our algorithm. The amount of time
needed for performing one specific shrink (grow) of $N_u$ ($N_s$) correlates
with the position of $N_u$ ($N_s$) on $K$; the closer $N_u$ ($N_s$) is to the
start (end) of $K$ the bigger amount of time needed for the shrink (grow) can
be expected.

Therefore, we propose to shrink (grow) only some of the local MUSes (MSSes)
based on their position on $K$. Let $|K|$ be the length of $K$, $u$ the index
of $N_u$ in $K$, and $S : \mathbb{N} \rightarrow \mathbb{N}$ be an arbitrary
user defined function. Our algorithm shrinks $N_u$ into a MUS if and only if $u
< S(|K|)$. As an example, consider $S(x) = \frac{x}{2}$; in such case $N_u$ is
shrunk only if it is contained in the first half of $K$.
Similarly, let $s$ be the index of local MSS $N_s$ of chain $K$ and $G :
\mathbb{N} \rightarrow \mathbb{N}$. The local MSS $N_s$ is grown only if $s >
|K| - G(|K|)$, which for example for $G(x) = \frac{x}{2}$ means that $N_s$ is
grown only if it is contained in the second half of $K$.
The complexity of performing shrinks also depends on the type of constrained
system that is being processed, therefore the concrete choice of $S$ and $G$ is
left as a parameter of our algorithm. 
Algorithm~\ref{alg-process-chain} shows an extended version of the method
$processChain$ which is able to produces MUSes and MSSes during its execution
based on the above mechanism.

\subsection{Example Execution of Our Algorithm}

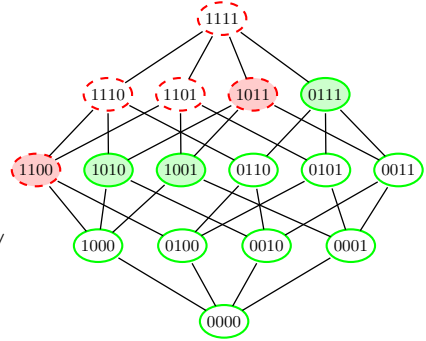
\begin{figure}[H]
\noindent
\begin{minipage}{0.55\textwidth}
\textbf{I. iteration}\\
-- Unex. couple $\langle 0000, 1111 \rangle$\\
-- Unex. chain $\langle 0000, 1000, 1100, 1110, 1111 \rangle$\\
-- A local MSS $1000$ and local MUS $1100$\\ are found\\
-- $1000$ is grown to the MSS $1010$\\
-- $1100$ is shrunk to the MUS $1100$\\
-- $MSS_{can} = \emptyset$ is updated to $\{1010\}$\\
-- $MUS_{can} = \emptyset$ is updated to $\{1100\}$\\
-- $f(Unex)$ is set to $(x_2 \vee x_4) \wedge (\neg x_1 \vee \neg x_2)$\\
\end{minipage}
\begin{minipage}{0.45\textwidth}

\begin{tikzpicture}[-,>=stealth', shorten >=1pt,auto,node distance=1.6cm,minimum size=0.62cm,
  thick,every node/.style={draw,white,text=black,ellipse,inner sep=0, outer sep=0, scale=0.7},
  sat/.style={draw,green,text=black},
  unsat/.style={dashed,red,text=black},
  mus/.style={fill=red!20},
  mss/.style={fill=green!20}]

\node[sat] (0) {$0000$};

\node[] (2) [above left = 0.7cm and 0.10cm of 0] {$0100$};
\node[sat] (1) [left of=2] {$1000$};
\node[sat] (3) [right of=2] {$0010$};
\node[] (4) [right of=3] {$0001$};

\node[unsat,mus] (12) [above left = 0.7cm and 0.37cm of 1] {$1100$};
\node[sat,mss] (13) [right = 0.32cm of 12] {$1010$};
\node[] (14) [right = 0.32cm of 13] {$1001$};
\node[] (23) [right = 0.32cm of 14] {$0110$};
\node[] (24) [right = 0.32cm of 23] {$0101$};
\node[] (34) [right = 0.32cm of 24] {$0011$};

\node[unsat] (123) [above right = 0.7cm and 0.5cm of 12] {$1110$};
\node[unsat] (124) [right = 0.32cm of 123] {$1101$};
\node[] (134) [right = 0.32cm of 124] {$1011$};
\node[] (234) [right = 0.32cm of 134] {$0111$};

\node[unsat] (1234) [above right = 0.7cm and 0.1cm of 124] {$1111$};

\path[every node/.style={font=\sffamily\small},line width=0.5pt]
(0) 	edge node [left,thin, dashed] {} (1)
		edge node [left] {} (2)
		edge node [left] {} (3)
		edge node [left] {} (4)
(1) 	edge node [left] {} (12)
		edge node [left] {} (13)
		edge node [left] {} (14)
(2) 	edge node [left] {} (12)
		edge node [left] {} (23)
		edge node [left] {} (24)
(3) 	edge node [left] {} (13)
		edge node [left] {} (23)
		edge node [left] {} (34)
(4) 	edge node [left] {} (14)
		edge node [left] {} (24)
		edge node [left] {} (34)
		
(12) 	edge node [left] {} (123)
		edge node [left] {} (124)
(13) 	edge node [left] {} (123)
		edge node [left] {} (134)
(14) 	edge node [left] {} (124)
		edge node [left] {} (134)
(23) 	edge node [left] {} (123)
		edge node [left] {} (234)	
(24) 	edge node [left] {} (124)
		edge node [left] {} (234)
(34) 	edge node [left] {} (134)
		edge node [left] {} (234)
		
(123) 	edge node [left] {} (1234)
(124) 	edge node [left] {} (1234)
(134) 	edge node [left] {} (1234)
(234) 	edge node [left] {} (1234)																	
;
\end{tikzpicture}
\end{minipage}
\vspace{10pt}

\noindent
\begin{minipage}{0.55\textwidth}
\textbf{II. iteration}\\
-- Unexplored couple $\langle 0001, 1011 \rangle$\\
-- Unexplored chain $\langle 0001, 1001, 1011 \rangle$\\
-- Local MSS $1001$, local MUS $1011$\\
-- $1001$ is grown to the MSS $1001$\\
-- $1011$ is shrunk to the MUS $1011$\\
-- $MSS_{can} \gets MSS_{can} \cup \{1001\}$\\
-- $MUS_{can} \gets MUS_{can} \cup \{1011\}$\\
-- $f(Unex) \equiv (x_2 \vee x_4) \wedge (x_2 \vee x_3) \wedge (\neg x_1 \vee \neg x_2) \wedge (\neg x_1 \vee \neg x_3 \vee \neg x_4)$\\
\end{minipage}
\begin{minipage}{0.45\textwidth}

\begin{tikzpicture}[-,>=stealth', shorten >=1pt,auto,node distance=1.6cm,minimum size=0.62cm,
  thick,every node/.style={draw,white,text=black,ellipse,inner sep=0, outer sep=0, scale=0.7},
  sat/.style={draw,green,text=black},
  unsat/.style={dashed,red,text=black},
  mus/.style={fill=red!20},
  mss/.style={fill=green!20}]

\node[sat] (0) {$0000$};

\node[] (2) [above left = 0.7cm and 0.1cm of 0] {$0100$};
\node[sat] (1) [left of=2] {$1000$};
\node[sat] (3) [right of=2] {$0010$};
\node[sat] (4) [right of=3] {$0001$};

\node[unsat,mus] (12) [above left = 0.7cm and 0.37cm of 1] {$1100$};
\node[sat,mss] (13) [right = 0.32cm of 12] {$1010$};
\node[sat,mss] (14) [right = 0.32cm of 13] {$1001$};
\node[] (23) [right = 0.32cm of 14] {$0110$};
\node[] (24) [right = 0.32cm of 23] {$0101$};
\node[] (34) [right = 0.32cm of 24] {$0011$};

\node[unsat] (123) [above right = 0.7cm and 0.5cm of 12] {$1110$};
\node[unsat] (124) [right = 0.32cm of 123] {$1101$};
\node[unsat,mus] (134) [right = 0.32cm of 124] {$1011$};
\node[] (234) [right = 0.32cm of 134] {$0111$};

\node[unsat] (1234) [above right = 0.7cm and 0.1cm of 124] {$1111$};

\path[every node/.style={font=\sffamily\small},line width=0.5pt]
(0) 	edge node [left,thin, dashed] {} (1)
		edge node [left] {} (2)
		edge node [left] {} (3)
		edge node [left] {} (4)
(1) 	edge node [left] {} (12)
		edge node [left] {} (13)
		edge node [left] {} (14)
(2) 	edge node [left] {} (12)
		edge node [left] {} (23)
		edge node [left] {} (24)
(3) 	edge node [left] {} (13)
		edge node [left] {} (23)
		edge node [left] {} (34)
(4) 	edge node [left] {} (14)
		edge node [left] {} (24)
		edge node [left] {} (34)
		
(12) 	edge node [left] {} (123)
		edge node [left] {} (124)
(13) 	edge node [left] {} (123)
		edge node [left] {} (134)
(14) 	edge node [left] {} (124)
		edge node [left] {} (134)
(23) 	edge node [left] {} (123)
		edge node [left] {} (234)	
(24) 	edge node [left] {} (124)
		edge node [left] {} (234)
(34) 	edge node [left] {} (134)
		edge node [left] {} (234)
		
(123) 	edge node [left] {} (1234)
(124) 	edge node [left] {} (1234)
(134) 	edge node [left] {} (1234)
(234) 	edge node [left] {} (1234)																	
;
\end{tikzpicture}
\end{minipage}

\vspace{10pt}

\noindent
\begin{minipage}{0.55\textwidth}
\textbf{III. iteration}\\
-- Unexplored couple $\langle 0011, 0111 \rangle$\\
-- Unexplored chain $\langle 0011, 0111 \rangle$\\
-- Local MSS $0111$, local MUS $undefined$\\
-- $0111$ is grown to the MSS $0111$\\
-- $MSS_{can} \gets MSS_{can} \cup \{0111\}$\\
-- $f(Unex) \equiv (x_2 \vee x_4) \wedge (x_2 \vee x_3) \wedge (x_1) \wedge (\neg x_1 \vee \neg x_2) \wedge (\neg x_1 \vee \neg x_3 \vee \neg x_4)$
\end{minipage}
\begin{minipage}{0.45\textwidth}

\begin{tikzpicture}[-,>=stealth', shorten >=1pt,auto,node distance=1.6cm,minimum size=0.62cm,
  thick,every node/.style={draw,white,text=black,ellipse,inner sep=0, outer sep=0, scale=0.7},
  sat/.style={draw,green,text=black},
  unsat/.style={dashed,red,text=black},
  mus/.style={fill=red!20},
  mss/.style={fill=green!20}]

\node[sat] (0) {$0000$};

\node[sat] (2) [above left = 0.7cm and 0.1cm of 0] {$0100$};
\node[sat] (1) [left of=2] {$1000$};
\node[sat] (3) [right of=2] {$0010$};
\node[sat] (4) [right of=3] {$0001$};

\node[unsat,mus] (12) [above left = 0.7cm and 0.37cm of 1] {$1100$};
\node[sat,mss] (13) [right = 0.32cm of 12] {$1010$};
\node[sat,mss] (14) [right = 0.32cm of 13] {$1001$};
\node[sat] (23) [right = 0.32cm of 14] {$0110$};
\node[sat] (24) [right = 0.32cm of 23] {$0101$};
\node[sat] (34) [right = 0.32cm of 24] {$0011$};

\node[unsat] (123) [above right = 0.7cm and 0.5cm of 12] {$1110$};
\node[unsat] (124) [right = 0.32cm of 123] {$1101$};
\node[unsat,mus] (134) [right = 0.32cm of 124] {$1011$};
\node[sat,mss] (234) [right = 0.32cm of 134] {$0111$};

\node[unsat] (1234) [above right = 0.7cm and 0.1cm of 124] {$1111$};

\path[every node/.style={font=\sffamily\small},line width=0.5pt]
(0) 	edge node [left,thin, dashed] {} (1)
		edge node [left] {} (2)
		edge node [left] {} (3)
		edge node [left] {} (4)
(1) 	edge node [left] {} (12)
		edge node [left] {} (13)
		edge node [left] {} (14)
(2) 	edge node [left] {} (12)
		edge node [left] {} (23)
		edge node [left] {} (24)
(3) 	edge node [left] {} (13)
		edge node [left] {} (23)
		edge node [left] {} (34)
(4) 	edge node [left] {} (14)
		edge node [left] {} (24)
		edge node [left] {} (34)
		
(12) 	edge node [left] {} (123)
		edge node [left] {} (124)
(13) 	edge node [left] {} (123)
		edge node [left] {} (134)
(14) 	edge node [left] {} (124)
		edge node [left] {} (134)
(23) 	edge node [left] {} (123)
		edge node [left] {} (234)	
(24) 	edge node [left] {} (124)
		edge node [left] {} (234)
(34) 	edge node [left] {} (134)
		edge node [left] {} (234)
		
(123) 	edge node [left] {} (1234)
(124) 	edge node [left] {} (1234)
(134) 	edge node [left] {} (1234)
(234) 	edge node [left] {} (1234)																	
;
\end{tikzpicture}
\end{minipage}

\caption{An~example execution of our algorithm}
\label{fig:exec}
\end{figure}

The following example explains the execution of our algorithm on a simple
set of constraints. The example is illustrated in Fig.~\ref{fig:exec}.
Let $C = \{c_1 = a, \, c_2 = \neg a, \, c_3 = b, \, c_4 = \neg a \vee \neg b\}$,
$S(x) = x$, $G(x) = x$.

Initially $MSS_{can} = \emptyset$, $MUS_{can} = \emptyset$ and all nodes are
unexplored, i.e. $f(Unex) = True$. Figure~\ref{fig:exec} shows the values of
control variables in each iteration and also illustrates the current states of
$\mathcal{P}(C)$. In order to save space we encode nodes as bitvectors,
for example the node $\{c_1,c_3,c_4\}$ is written as $1011$.

After the last iteration of the first phase of our algorithm there is no model
of $f(Unex)$ (this means that $Unex$ is empty), $MSS_{can} = \{1010, 1001,
0111\}$ and $MUS_{can} = \{1100, 1011\}$. Because functions $S$ and $G$ were
stated in this example as $S(x) = x, G(x) = x$, each candidate on MUS or MSS
has been alredy shrunk or grown to MUS or MSS, respectively, therefore $MSS(C)
= MSS_{can}, MUS(C) = MUS_{can}$ and the second phase of our algorithm can be
omitted.

Note that in the first iteration the node $1010$ was found to be a MSS, which
means (due to the definition of MSS) that all its supersets are unsatisfiable.
One would use this fact to mark all supersets of $1010$ as explored, however our
algorithm does not do this because some of these subsets can be MUSes ($1011$
in this example). If we were interested only in MSS enumeration we could mark
all supersets of each MSS as explored; dually in the case of only MUS
enumeration.

\section{Experimental Results}\label{sec:exp}

We now demonstrate the performance of several
variants of our algorithm on a variety of Boolean CNF benchmarks. In
particular, we implemented in C++ both the Basic and the Pivot Based approach
for constructing chains and we evaluated both these approaches using several
variants of the functions $S$ and~$G$. We also give a comparison with MARCO
algorithm~\cite{marco}.

MARCO algorithm was presented by its authors in two variants, the basic
variant and the optimised variant which is tailored for MUS enumeration. Both
variants are iterative. The basic variant finds in each iteration 
an~unexplored node, checks its satisfiability and based on the result the node is
either shrunk into a MUS or grown into a MSS. Subsequently, MARCO uses
the monotonicity of $\mathcal{P}(C)$ to deduce satisfiability of other nodes in
the same way our algorithm does.
The optimised variant differs from the basic variant in the selection of the
unexplored node; it always selects a~maximal unexplored node. If the node is
unsatisfiable it is shrunk into a MUS, otherwise it is guaranteed to be a MSS.
We used the optimised variant in our experiments. The pseudocodes of both
variants can be found in~\cite{marco}. 

Note that both compared algorithms (MARCO and our algorithm) employ several
external tools during their execution, namely a~SAT solver for finding the
unexplored nodes, a~constraint solver to decide the satisfiability of
constraint sets, and the two procedures \emph{shrink} and \emph{grow} mentioned
above. The list of external tools coincides for both algorithms.
Therefore, we reimplemented MARCO algorithm in C++ to ensure that the
two algorithms use the same implementations of the shrink and grow methods
and the same solvers. As both the SAT solver and constraint solver we used
the miniSAT tool~\cite{minisat} and we used the simple implementation of the shrink and grow
methods as described earlier. Note that there are some efficient
implementations of the shrink and grow methods for Boolean constraints,
however, in general there might be no effective implementation these methods.
That is why we used the simple implementations.

\begin{table}[t]
\caption{The number of instances in which the algorithms output at least one
MSS (the first number in each cell) or MUS (the second number).}
  \label{res:at-least-one}

  \medskip
\centering
\scalebox{0.8}{
  \begin{tabular}{ c c | c | c | c | c | c | c}
	& \diagbox{$G(x)$}{$S(x)$}	& $x$ & $0.8x$ & $0.6x$ & $0.4x$ & $0.2x$ & $0x$\\ \cline{2-8}
	\parbox[t]{2mm}{\multirow{6}{*}{\rotatebox[origin=c]{90}{Basic approach}}} & $x$    & 56 $|$ 56 & 151 $|$ 40 & 150 $|$ 33 & 144 $|$ 12 & 149 $|$ 16 & 151 $|$ 0 \\ \cline{2-8}
	& $0.8x$ & 56 $|$ \textbf{60} & 149 $|$ 44 & 151 $|$ 37 & 144 $|$ 16 & 150 $|$ 20 & \textbf{152} $|$ 0 \\ \cline{2-8}
	& $0.6x$ & 56 $|$ \textbf{60} & 149 $|$ 44 & 144 $|$ 35 & 144 $|$ 18 & 151 $|$ 22 & 151 $|$ 0 \\ \cline{2-8}
	& $0.4x$ & 54 $|$ \textbf{60} & 149 $|$ 45 & 140 $|$ 36 & 143 $|$ 32 & 150 $|$ 30 & 151 $|$ 0 \\ \cline{2-8}
	& $0.2x$ & 53 $|$ \textbf{60} & 148 $|$ 45 & 138 $|$ 43 & 138 $|$ 40 & 144 $|$ 35 & 145 $|$ 0 \\ \cline{2-8}
	& $0x$ & 0 $|$ \textbf{60} & 0 $|$ 47 & 0 $|$ 46 & 0 $|$ 44 & - $|$ - & 0 $|$ 0 \\ \hhline{=|=|=|=|=|=|=|=}
	\parbox[t]{2mm}{\multirow{6}{*}{\rotatebox[origin=c]{90}{Pivot based approach}}} & $x$    & 56 $|$ 56 & 151 $|$ 40 & 151 $|$ 32 & 151 $|$ 14 & 151 $|$ 12 & 144 $|$ 0 \\ \cline{2-8}
	& $0.8x$ & 56 $|$ \textbf{60} & 151 $|$ 43 & 151 $|$ 36 & 150 $|$ 18 & 149 $|$ 16 & 145 $|$ 0 \\ \cline{2-8}
	& $0.6x$ & 56 $|$ \textbf{60} & 151 $|$ 43 & 151 $|$ 35 & 151 $|$ 18 & \textbf{152} $|$ 16 & 144 $|$ 0 \\ \cline{2-8}
	& $0.4x$ & 54 $|$ \textbf{60} & 150 $|$ 43 & 147 $|$ 35 & 151 $|$ 14 & 150 $|$ 13 & 144 $|$ 0 \\ \cline{2-8}
	& $0.2x$ & 51 $|$ \textbf{60} & 146 $|$ 45 & 145 $|$ 31 & 148 $|$ 12 & 148 $|$ 12 & 143 $|$ 0 \\ \cline{2-8}
	& $0$ & - $|$ - & - $|$ - & - $|$ - & - $|$ - & 0 $|$ 0 \\ \hhline{=|=|=|=|=|=|=|=}
	& MARCO  & \textbf{51} $|$ \textbf{51}
  \end{tabular}
}
\end{table}

As an experimental data we used a collection of 294 unsatisfiable Boolean CNF
Benchmarks that were taken from the MUS track of the 2011 SAT
competition\footnote{\url{http://www.cril.univ-artois.fr/SAT11/}}.
The benchmarks range in their size from 70 to 16 million constraints and from
26 to 4.4 million variables and were drawn from a variety of domains and
applications. All experiments were run with a time limit of 60 seconds.

Due to the potentially exponentially many MUSes and/or MSSes in each instance,
the complete MUS and MSS enumeration is generally intractable. Moreover, even
outputting a single MUS/MSS can be intractable for larger instances as it
naturally includes solving the satisfiability problem, which is for Boolean
instances NP-complete. Table~\ref{res:at-least-one} shows in how many instances
the variants of our algorithm were able to output at least one MUS or MSS.
MARCO was able to output at least one MUS and one MSS in 51 instances whereas
several variants of our algorithm were able to output some MSSes in about 150
instances and some MUSes in up to 60 instances. Some of the 296 instances are
just intractable for the solver which is not able to perform even a single
consistency check within the used time limit.
The other significant factor that affected the results is the complexity of the shrink method. MARCO in every iteration either "hits" a satisfiable node and directly outputs it as an MSS or waits till the shrink method shrinks the unsatisfiable node into a MUS. Therefore, each call of the shrink method can suspend the execution for a nontrivial time.

One can see that our algorithm also suffers from the possibly very expensive
shrink calls and performs very poorly when the $S$ function is set to $S(x) = x$. 
On the other hand, the variants that perform only the ``easier'' shrinks by
setting $S$ to be $S(x) < x$ achieved better results. 
The grow method is generally cheaper to perform than the shrink method as 
checking whether an addition of a constraint to a satisfiable set of
constraints makes this set unsatisfiable is usually cheaper than the dual task.
No significant difference between the Basic and the Pivot based approach was
captured in this comparison.

\begin{table}[t]
\caption{The 5\% trimmed sum of outputted MSSes and MUSes (summed over all 294
instances). The first number in each cell is the number of outputted MSSes, the
second is the number of outputted MUSes.}

\medskip
\centering
\scalebox{0.8}{
  \begin{tabular}{ c c | c | c | c | c | c | c}
	  & \diagbox[width=2.5cm]{$G(x)$}{$S(x)$}	& $x$ & $0.8x$ & $0.6x$ & $0.4x$ & $0.2x$ & $0x$\\ \cline{2-8}
	\parbox[t]{2mm}{\multirow{6}{*}{\rotatebox[origin=c]{90}{Basic approach}}} & $x$ & 1744 $|$ 339 & 9798 $|$ 212 & 9936 $|$ 87 & 6942 $|$ 0 & 9726 $|$ 2 & 10216 $|$ 0 \\ \cline{2-8}
	& $0.8x$ & 1741 $|$ 344 & 9908 $|$ 217 & 9756 $|$ 94 & 6787 $|$ 2 & 9684 $|$ 6 &  9378 $|$ 0 \\ \cline{2-8}
	& $0.6x$ & 1740 $|$ 348 & 9859 $|$ 224 & 6969 $|$ 40 & 6999 $|$ 4 & 9696 $|$ 8 &  9436 $|$ 0 \\ \cline{2-8}
	& $0.4x$ & 1877 $|$ 436 & 10013 $|$ 252 & 7218 $|$ 67 & 7694 $|$ 50 & 10420 $|$ 39 &  10114 $|$ 0 \\ \cline{2-8}
	& $0.2x$ & 1757 $|$ \textbf{635} & 10161 $|$ 527 & 7925 $|$ 262 & 8196 $|$ 101 & \textbf{10853} $|$ 66 & 10111 $|$ 0 \\ \cline{2-8}
	& $0$ & 0 $|$ 632 & 0 $|$ 554 & 0 $|$ 356 & 0 $|$ 107 & - $|$ - & 0 $|$ 0 \\ \hhline{=|=|=|=|=|=|=|=}
	\parbox[t]{2mm}{\multirow{6}{*}{\rotatebox[origin=c]{90}{Pivot based approach}}} & $x$    & 2535 $|$ 349 & 8330 $|$ 208 & 7775 $|$ 71 & 6705 $|$ 0 & 6725 $|$ 0 & 5089 $|$ 0 \\ \cline{2-8}
	& $0.8x$ & 2660 $|$ 492 & 8336 $|$ 255 & 7680 $|$ 85 & 6961 $|$ 4 & 6889 $|$ 2 &  5061 $|$ 0 \\ \cline{2-8}
	& $0.6x$ & 2771 $|$ 567 & 8481 $|$ 290 & 7779 $|$ 92 & 7066 $|$ 4 & 6830 $|$ 2 &  5067 $|$ 0 \\ \cline{2-8}
	& $0.4x$ & 2814 $|$ 597 & 8418 $|$ 388 & 7975 $|$ 145 & 6814 $|$ 0 & 6950 $|$ 0 &  5302 $|$ 0 \\ \cline{2-8}
	& $0.2x$ & 2763 $|$ \textbf{837} & \textbf{8633} $|$ 697 & 7220 $|$ 41 & 6563 $|$ 0 & 6409 $|$ 0 & 4910 $|$ 0 \\ \cline{2-8}
	& $0$ & - $|$ - & - $|$ - & - $|$ - & - $|$ - & - $|$ - & 0 $|$ 0 \\ \hhline{=|=|=|=|=|=|=|=}
	& MARCO  & \textbf{749} $|$ \textbf{215}
  \end{tabular}
}
\label{res:sum}  
\end{table}

Another comparison can be found in Table~\ref{res:sum} that shows the 5\%
trimmed sums of outputted MSSes and MUSes (summed over all of the 294
instances), i.e. 5\% of the instances with the least outputted MSSes (MSSes)
and 5\% of the instances with the most outputted MSSes (MSSes) were discarded.
All variants of our algorithm were noticeably better in MSS enumeration than
MARCO. In the case of MUS enumeration MARCO outperformed these variants
of our algorithm that shrink only some of the local MUSes, i.e. variants where
$S(x) = 0.6x$ and $S(x) = 0.4x$. However, the variants with $S(x) = x$ and $S(x)
= 0.8x$ performed better, especially the variant with $G(x) = 0.2x, \, S(x) = x$
outputted about three times more MUSes than  MARCO. In this comparison, there
is already some notable difference between the Basic and the Pivot based
approach. The Pivot based approach seems to be better for MUS enumeration
whereas the Basic approach is more suitable for the MSS enumeration. As the Pivot
based approach is randomized its performance may vary if it is run repeatedly
on the same instances; result of a single run may be misleading. Therefore, we
ran all tests of the Pivot based approach repeatedly and the tables show the
average values.

\begin{figure}[t]
\begin{minipage}{0.5\textwidth}
  \includegraphics[scale=0.5]{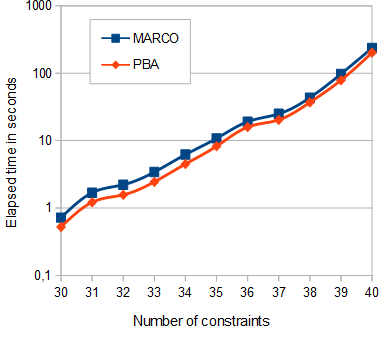}
\end{minipage}
\begin{minipage}{0.5\textwidth}
  \includegraphics[scale=0.5]{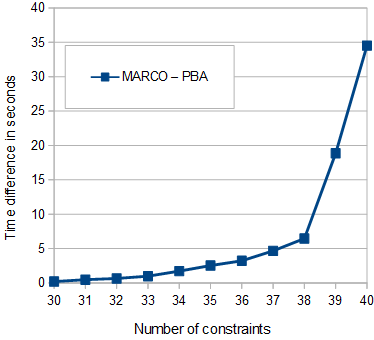}
\end{minipage}

\caption{The time comparison of MARCO and our algorithm. The chart on the left side shows the running times and it is logarithmically scaled. The other chart shows the difference in the running times between the two algorithms.}
\label{res:time}
\end{figure}

Besides the number of outputted MUSes/MSSes within a given time limit, we also
compared our algorithm with MARCO in the case of complete MUS/MSS
enumeration. We used the generator of Boolean CNF formulae from~\cite{cnfgen}
to generate tractable instances with a size of 30 to 40 constraints, 15
instances per each size. 
The graphs in Fig.~\ref{res:time} show the time comparison of MARCO and our
algorithm using the Pivot based approach with $S$ and $G$ set to $S(x) = 0.2x$
and $G(x) = 0.8x$. All of the instances were tractable which means that both
phases of our algorithm were executed. Some of the MUSes and MSSes were output
in the online manner, the rest of them were extracted from the candidate sets
in the second phase.

Summarised, our algorithm outperformed MARCO both in the online MUS and MSS
enumeration and in the complete MUS and MSS enumeration. 
Also, the results show that the choice of the functions $S$ and $G$ greatly
affect the efficiency of our algorithm. The faster the $S$ ($G$) grows, the
more effort is made to output MUSes (MSSes). Also, it may be worth to
always perform at least the ``easy'' grows (shrinks) even if we want to output
only MUSes (MSSes), because each shrink (grow) also helps to reduce the space
of unexplored nodes.

\section{Conclusion}\label{sec:concl}
In this paper, we have presented a novel algorithm for online enumeration of MUSes and MSSes which is applicable to any type of constraint system. 
The core of the algorithm is based on a novel binary-search-based approach which allows the algorithm to efficiently explore the space of all subsets of a given set of constraints.
We have made an experimental comparison with MARCO, the state-of-the-art algorithm for online MUS and MSS enumeration. The results show that our algorithm is better both for online enumeration and also in the case of complete enumeration. Our algorithm can be built on a top of any consistency solver and can employ any implementation of the \emph{shrink} and \emph{grow} methods, therefore any future advance in this areas can be reflected in the performance of our algorithm.

One direction of future research is to aim at parallel processing of the search space in order to improve the performance of our approach; there are usually many disjoint unexplored chains that can be processed concurrently.
Another possible direction is to focus on some specific types of constraint systems and customise our algorithm to be more efficient for these systems.

\bibliographystyle{splncs03}
\bibliography{main}

\end{document}